\documentclass[10pt,twocolumn,letterpaper]{article}

\usepackage[pagenumbers]{cvpr} 
\definecolor{cvprblue}{rgb}{0.21,0.49,0.74}
\usepackage[pagebackref,breaklinks,colorlinks,allcolors=cvprblue]{hyperref}

\usepackage{algorithm}
\usepackage{algorithmic}
\usepackage{graphicx}
\usepackage{amsmath}

\usepackage{float}
\usepackage{mathtools}
\usepackage{algorithm}
\usepackage{algorithmic}
\usepackage{enumitem}
\usepackage{lipsum}
\usepackage{booktabs}
\usepackage[table,xcdraw]{xcolor}
\usepackage[normalem]{ulem}
\useunder{\uline}{\ul}{}
\usepackage{ulem} 

\def\paperID{17847} 
\def\confName{CVPR}
\def\confYear{2026}

\title{BeyondFacial: Identity-Preserving Personalized Generation \\Beyond Facial Close-ups
}

\author{
    Songsong Zhang\textsuperscript{1†}, Chuanqi Tang\textsuperscript{2†}, Hongguang Zhang\textsuperscript{3}, Minglong Li\textsuperscript{1}, Xueqiong Li\textsuperscript{1},\\ 
    Shaowu Yang\textsuperscript{1},Yuanxi Peng\textsuperscript{1}, Wenjing Yang\textsuperscript{1}, Jing Zhao\textsuperscript{1}*\\
    {\small \textsuperscript{1}DIDS, NUDT, China,}
    {\small \textsuperscript{2}TJU, China,}
    {\small \textsuperscript{3}Unit 32010 of the Chinese PLA} \\
    {\tt\small zhaojing@nudt.edu.cn}
}   
    
\begin{document}
\twocolumn[{%
\renewcommand\twocolumn[2][]{#1}%
\maketitle%
\centering \centering
\includegraphics[width=\textwidth]{figs/First_Page2.pdf}
\captionof{figure}{
Existing Identity-Preserving Personalized Generation (IPPG) methods over-rely on facial close-ups. Awkward backgrounds, incomplete characters, and truncated semantics hinder visual storytelling. Ours transcends this limitation, enabling holistic scenes and rich, full-fledged character generation beyond close-ups.
}

\vspace*{0.2cm}
\label{fig:first_page}}] 

\maketitle

\begin{abstract}

Identity-Preserving Personalized Generation (IPPG) has advanced film production and artistic creation, yet existing approaches overemphasize facial regions—resulting in outputs dominated by "facial close-ups." These methods suffer from weak visual narrativity and poor semantic consistency under complex text prompts, with the core limitation rooted in identity (ID) feature embeddings undermining the semantic expressiveness of generative models.  
To address these issues, this paper presents an IPPG method that breaks the constraint of facial close-ups, achieving synergistic optimization of identity fidelity and scene semantic creation. Specifically, we design a Dual-Line Inference (DLI) pipeline with identity-semantic separation, resolving the representation conflict between ID and semantics inherent in traditional single-path architectures. Further, we propose an Identity Adaptive Fusion (IdAF) strategy that defers ID-semantic fusion to the noise prediction stage, integrating adaptive attention fusion and noise decision masking to avoid ID embedding interference on semantics without manual masking. Finally, an Identity Aggregation Prepending (IdAP) module is introduced to aggregate ID information and replace random initializations, further enhancing identity preservation.  
Experimental results validate that our method achieves stable and effective performance in IPPG tasks beyond facial close-ups, enabling efficient generation without manual masking or fine-tuning. As a plug-and-play component, it can be rapidly deployed in existing IPPG frameworks, addressing the over-reliance on facial close-ups, facilitating film-level character-scene creation, and providing richer personalized generation capabilities for related domains.

\end{abstract}

\section{Introduction}
\label{sec:intro}

The rapid advancement of Identity-Preserving Personalized Generation (IPPG)~\cite{magicnaming,ipadapter,PhotoMaker,PuLID,InstantID,facestudio,FastComposer} has reshaped film production and artistic creation. By precisely anchoring target identity features and fusing text prompts for personalized synthesis, IPPG has become a core tool for boosting creative efficiency and expanding creative boundaries. 

\begin{figure}[t]
  \centering
   \includegraphics[width=\linewidth]{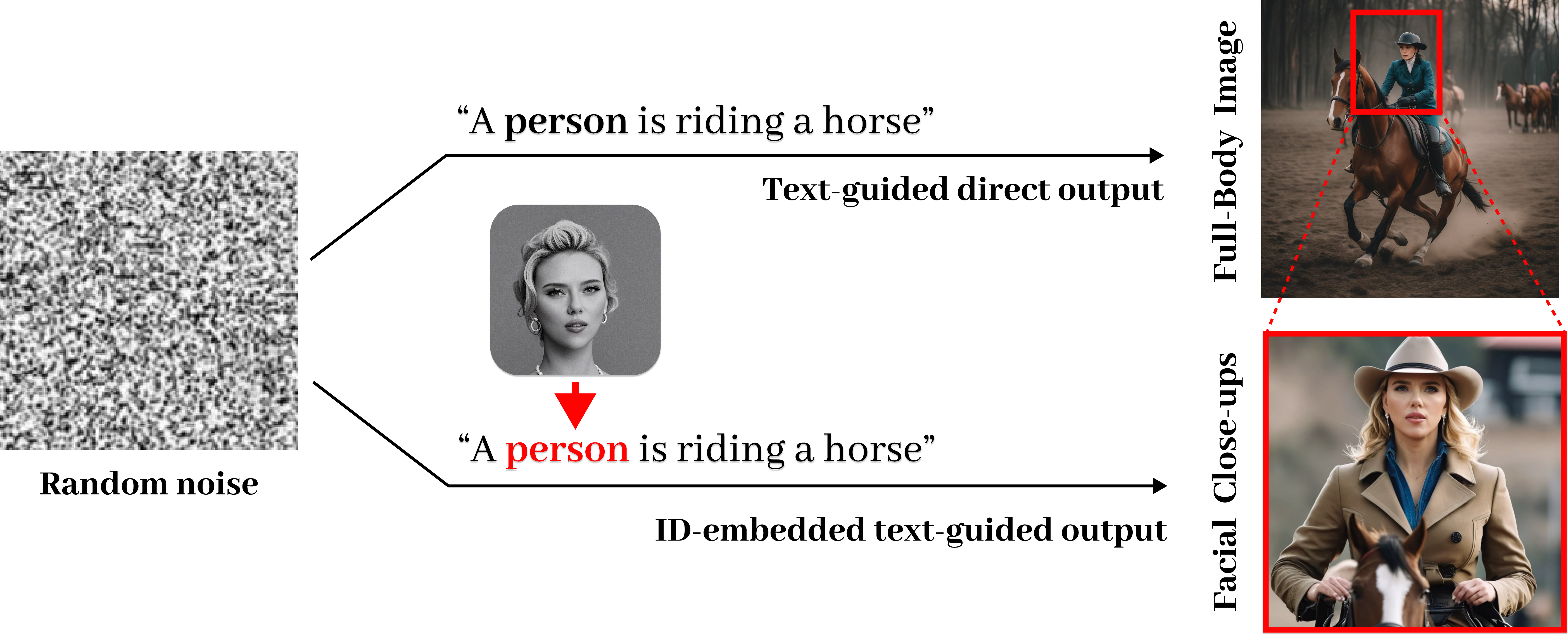}
   \caption{Under the same random seed, ID embedding transforms full-body images into facial close-ups. }
   \label{fig:intro}
\end{figure}

However, current IPPG research faces critical bottlenecks: existing methods overemphasize facial regions, resulting in outputs dominated by``facial close-ups." This not only degrades visual narrativity—manifested in incomplete character depictions and abrupt background transitions—but also causes severe semantic inconsistency under complex text prompts, including scale misalignment, illogical layouts, and generative artifacts. Through systematic analysis (Fig.\ref{fig:intro}), we identify the root cause: the manner in which identity (ID) features are embedded in traditional frameworks directly undermines the inherent semantic expressiveness of generative models. This creates an intractable conflict between ID fidelity and scene semantic creativity, ultimately leading to the prevalence of facial close-ups.  

To break this constraint, we propose a novel IPPG method that achieves synergistic optimization of ID preservation and scene semantic creation. Addressing the representational conflict between ID and semantics in traditional single-path architectures, we design a Dual-Line Inference (DLI) pipeline with identity-semantic separation. It employs an independent ID branch to ensure identity consistency and a dedicated semantic branch to construct complete scenes, resolving the competition for representational space at the architectural level. Building on DLI, we introduce the Identity Adaptive Fusion (IdAF) strategy: it defers ID-semantic fusion to the noise prediction stage, integrating adaptive attention fusion and a noise decision masking mechanism. This eliminates ID embedding interference on semantic expression \textit{without manual masking}. To further enhance identity fidelity, we propose the Identity Aggregation Prepending (IdAP) module, which aggregates ID information and replaces randomly initialized ID features, effectively suppressing interference from irrelevant identities during semantic generation.  

Experimental results demonstrate that our method achieves stable and efficient performance in IPPG tasks beyond facial close-ups. It enables high-quality generation without manual masking or model fine-tuning and can be rapidly deployed as a plug-and-play component in existing IPPG frameworks. By resolving the long-standing``facial close-up" limitation, our work provides robust technical support for film-level character-scene co-creation, significantly expanding IPPG’s application scope in creative fields. \textbf{Our contributions can be summarized as follows:}
\begin{itemize}

\item We proposed a Dual-Line Inference (DLI) pipeline with identity-semantic separation, resolving the ID-semantic representational conflict in traditional single-path architectures to enable non-close-up IPPG.  

\item We designed an Identity Adaptive Fusion (IdAF) strategy, deferring fusion to noise prediction with adaptive attention and noise masking for interference-free ID-semantic synergy, no manual masking needed.  

\item We introduced an Identity Aggregation Prepending (IdAP) module, aggregating ID info and replacing random initialization to suppress irrelevant identity interference, boosting ID fidelity in complex scenes.  

\item We presented a lightweight plug-and-play solution, integrable into existing IPPG frameworks without fine-tuning, eliminating ``facial close-up" limitations and expanding applications.
\end{itemize}

\section{Related Works}
\label{sec:related_works}

\begin{figure*}[t]
  \centering
   \includegraphics[width=\linewidth]{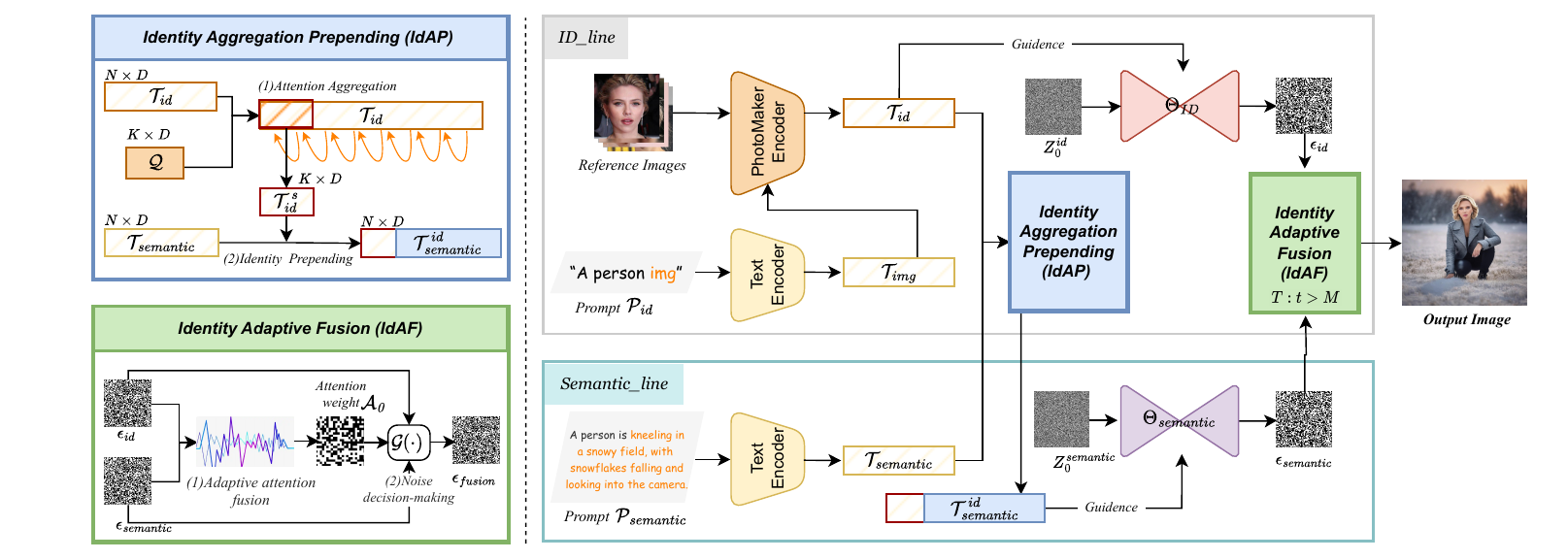}
   \caption{Overview of framework, integrating three key innovations: (1) Dual-Line Inference (DLI) pipeline, separating identity and semantic streams to resolve their representational conflict; (2) Identity Adaptive Fusion (IdAF) strategy, deferring fusion to noise prediction with adaptive attention/masking for interference-free fusion without manual input; (3) Identity Aggregation Prepending (IdAP) module, aggregating ID info and replacing random initializations to enhance preservation. This framework enables high-quality IPPG beyond facial close-ups via identity-semantic harmonization.}
   \label{fig:method}
\end{figure*}

\textbf{Personalized Generation based on Diffusion Models.}  
Advances in diffusion models~\cite{Arar_Gal_Atzmon_Chechik_Cohen-Or_Shamir_Bermano_2023,Chen_Zhang_Wang_Duan_Zhou_Zhu_2023,Gal_Arar_Atzmon_Bermano_Chechik_Cohen-Or_2023,Hinz_Heinrich_Wermter_2022,Li_Wen_Shi_Yang_Huang_2022,Liu_Chilton_2022,Liu_Zhang_Ma_Peng_Liu_2023,Prabhudesai_Goyal_Pathak_Fragkiadaki_2023,Shan_Ding_Passananti_Zheng_Zhao_2023,Xie_Li_Huang_Liu_Zhang_Zheng_Shou_2023,nulltextcartoon} have established personalized generation~\cite{gal2022image,dreambooth22,Wei_Zhang_Ji_Bai_Zhang_Zuo_2023,Hua_Liu_Ding_Liu_Wu_He_2023,NeTI,Ku_Li_Zhang_Lu_Fu_Zhuang_Chen_2023} as a cornerstone of generative modeling, enabling synthesis of customized content for specific objects, styles, or concepts while preserving their unique traits across contexts. DreamBooth~\cite{DreamBooth} marked a milestone, with subsequent work~\cite{Textual_Inversion,InstantBooth,Wei_Zhang_Ji_Bai_Zhang_Zuo_2023,Multi_Concept,magicfusion} addressing key limitations like high computational cost and overfitting. Textual Inversion~\cite{Textual_Inversion} introduced parameter efficiency by optimizing only target text embeddings (no full model tuning) for faster deployment. MagicFusion~\cite{magicfusion} integrated reference targets into scene semantics via saliency-based noise fusion, enhancing consistency with complex text prompts. Custom Diffusion~\cite{Multi_Concept} enabled multi-concept generation through co-training or closed-form merging of fine-tuned models. InstantBooth~\cite{InstantBooth} achieved instantaneous text-guided personalization using pre-trained models, eliminating test-time finetuning.

\noindent \textbf{Identity-Preserving Generation.}  
Identity-preserving personalized image generation~\cite{facestudio,FastComposer,PhotoMaker,PortraitBooth,InstantID,Zhang_Qi_Zhang_Zhang_Wu_Chen_Chen_Wang_Wen_2022,Pikoulis_Filntisis_Maragos_2023,magicnaming,PuLID,ipadapter} focuses on generating personalized images centered on a specific individual, with identity consistency as a critical metric.  Early methods required per-identity fine-tuning at inference~\cite{dreambooth22,SVDiff}. Subsequent tuning-free methods~\cite{Face0,Inserting-Anybody} emerged, encoding identity information directly into the generation pipeline. However, while prioritizing high identity fidelity, these approaches significantly degrade the base T2I model's original capabilities.
MagicNaming~\cite{magicnaming} learns an ID namer from Laion5B to predict``names" for facial images, enabling consistent generation for ordinary individuals. IP-Adapter~\cite{ipadapter} supports style-transferred generation from prompts without LoRA training, accommodating multiple reference images and feature extractions. InstantID~\cite{InstantID} guides generation via facial/landmark images and text prompts, using semantic and spatial constraints. PhotoMaker~\cite{PhotoMaker} processes multiple same-ID images through an encoder to produce stacked ID embeddings, strengthening identity-centric feature extraction. PuLID~\cite{PuLID} introduces contrastive alignment and precise ID losses, minimizing disruption to the base model while ensuring high ID fidelity.  

Existing works struggle to go beyond facial close-ups or support complex text-guided scene generation, severely limiting visual storytelling and practical value. Our work proposes an IPPG method that transcends facial close-ups, enabling holistic scene construction and rich character generation.

\section{Methods}

To address these limitations, we first diagnose why existing IPPG methods over-rely on facial close-ups. We then introduce our dual-line inference (DLI) pipeline. Building on DLI, we detail core innovations—IdAF and IdAP—critical for scene semantics and identity consistency. Fig.\ref{fig:method} overviews our framework.

\subsection{Analysis of Facial Close-ups}
Within personalized content generation, a core focus is aligning outputs with references—particularly for human-centric tasks, where identity fidelity is critical. Existing methods extract identity features via ID encoders, which are embedded into semantic tokens or diffusion cross-attention layers to enforce identity consistency. Yet such methods frequently overemphasize facial close-ups, limiting artistic and practical utility due to compromised scene completeness.  
Using PhotoMaker~\cite{PhotoMaker} as a case study, we experimentally identified the root cause (Fig.~\ref{fig:intro}): without ID features, diffusion models—guided by semantics alone—generate high-quality panoramic scenes. In contrast, with ID features injected, semantic embeddings are disrupted, driving over-focus on facial regions and thus close-ups.  

\noindent \textbf{These findings motivate two key adjustments:} (1) refining the injection of ID features into semantic representations; (2) adjusting embedding timing—first leveraging raw semantics to guide diffusion in forming a panoramic layout, then integrating ID features.

\subsection{Dual-Line Inference (DLI)}  
Existing methods use a single inference pathway to jointly enforce identity fidelity and semantic consistency, but co-embedding identity and scene semantics induces unavoidable representation conflicts. To resolve this, we propose a dual-line pipeline with independent ID and semantic lines: the ID line ensures identity fidelity, while the semantic line handles scene construction. We instantiate this using PhotoMaker~\cite{PhotoMaker} as the baseline.  

In the ID line: Given reference images $I_{ref}$ and the ID placeholder prompt $\mathcal{P}_{id}$(eg. ``A person img"), features are processed via an image encoder and text encoder, respectively, to yield the ID token $\mathcal{T}_{id}$. This token directly guides the ID generator $\Theta_{id}$ to predict ID noise $\epsilon_{id}$ via classifier-free guidance (CFG):  
\begin{equation}\label{text_encoder}
    \mathcal{T}_{id} = \mathbb{E}_{\text{img}}(I_{ref},\mathbb{E}_{\text{text}}(\mathcal{P}_{id}))
\end{equation}  
\begin{equation}\label{noise_predict_id}
    \epsilon_{id}(x_t \mid \mathcal{T}_{id}) = \text{CFG}(\Theta_{id}, \mathcal{T}_{id})
\end{equation}  

In the semantic line: Target scene prompts $\mathcal{P}_{\text{semantic}}$ are encoded via the text encoder to generate the semantic token $\mathcal{T}_{\text{semantic}}$. Our proposed IdAP module fuses $\mathcal{T}_{id}$ and $\mathcal{T}_{\text{semantic}}$ to produce the aggregated semantic token $\mathcal{T}_{\text{semantic}}^{id}$, which then guides the semantic generator $\Theta_{\text{semantic}}$ to predict semantic noise $\epsilon_{\text{semantic}}$:  
\begin{equation}\label{agg_token}
    \mathcal{T}_{\text{semantic}}^{id} = \textbf{IdAP}(\mathcal{T}_{id}, \mathcal{T}_{\text{semantic}})
\end{equation}  
\begin{equation}\label{noise_predict_sem}
    \epsilon_{\text{semantic}}(x_t \mid \mathcal{T}_{\text{semantic}}^{id}) = \text{CFG}(\Theta_{\text{semantic}}, \mathcal{T}_{\text{semantic}}^{id})
\end{equation}  

For the two predicted noises at the same timestep, our IdAF strategy performs interference-free fusion to obtain the final noise $\epsilon_{fusion}$ by follows,
\begin{equation}
    \epsilon_{\text{fusion}} = \textbf{IdAF}(\epsilon_{id},\epsilon_{semantic})
\end{equation}
Notably, IdAF is only activated when the sampling timestep $t > M$.

\subsection{Identity Adaptive Fusion (IdAF)}  
Building on our earlier findings, we defer ID-semantic fusion to post-denoising to preserve global visual quality, and propose IdAF to enable interference-free ID-semantic fusion.

Specifically, using an existing ID-preserving method (e.g., PhotoMaker~\cite{PhotoMaker}), we first generate two noise maps: (1) ID-only noise $\epsilon_{id} \in \mathbb{R}^{N \times C \times H \times W}$, guided by the placeholder prompt $\mathcal{P}_{id}$ and reference images; (2) semantic scene noise $\epsilon_{\text{semantic}} \in \mathbb{R}^{N \times C \times H \times W}$ for arbitrary identities, guided by target semantic prompts$\mathcal{P}_{semantic}$.  

We construct an attention weight tensor $\mathcal{A}$ by stacking these noises along a decision dimension $\mathcal{D}$:  
\begin{equation}\label{eq:A}
    \mathcal{A} = [\epsilon_{id}, \epsilon_{\text{semantic}}]_{\mathcal{D}} \in \mathbb{R}^{2 \times N \times C \times H \times W}
\end{equation}  

After element-wise absolute value, reshaping, and resizing, we obtain the initial weight tensor $\mathcal{A}_0$:  
\begin{equation}\label{eq:A0}
    \mathcal{A}_0 = \mathcal{S}_2 \circ \mathcal{S}_1 \circ \mathcal{R}\left( |\mathcal{A}|, (-1, C_{\text{mid}}, H, W) \right)
\end{equation}  
where $|\cdot|$ denotes element-wise absolute value, $\mathcal{R}(\cdot, \text{dim})$ reshapes to dimension $\text{dim}$, $\mathcal{S}_1, \mathcal{S}_2$ are consecutive spatial resizing operations to align attention features with noise predictions, and $\circ$ denotes function composition.  

To align channels, we reshape $\mathcal{A}_0$ into a standard form with decision dimensions, yielding $\bar{\mathcal{A}}_0$:  
\begin{equation}
    \bar{\mathcal{A}}_0 = \mathcal{M}\left( \mathcal{R}(\mathcal{A}_0, (2, N, C_{\text{mid}}, H, W)), \text{dim}=2 \right)
\end{equation}  
where $\mathcal{M}(\cdot, \text{dim})$ computes the mean over dimension $\text{dim}$. We then spatially normalize the two decision tensors of $\bar{\mathcal{A}}_0$ and introduce scaling factors to enhance saliency:  
\begin{equation}\label{eq:A0i}
    \mathcal{A}_0^{(i)} = \mathcal{R}\left( \mathcal{SM}\left( \lambda^{(i)} \cdot \mathcal{R}(\bar{\mathcal{A}}_0^{(i)}, \text{dim}_1) \right), \text{dim}_2 \right)
\end{equation}  
with $\text{dim}_1 = (N \cdot C, H \cdot W)$, $\text{dim}_2 = (N, C, H, W)$, $i=0,1$, $\mathcal{SM}(\cdot)$ as the softmax function, and $\lambda^{(i)} > 0$ as weight scalers.  

The final fused noise is obtained by selecting elements from $\epsilon_{id}$ and $\epsilon_{\text{semantic}}$ using the optimal indices from $\mathcal{A}_0$ along $\mathcal{D}$:  
\begin{equation}
    \epsilon_{\text{fusion}} = \mathcal{G}\left( \mathcal{O}, \arg\max_{\mathcal{D}} \mathcal{A}_0, \mathcal{D} \right)
\end{equation}  
where $\mathcal{G}(\cdot, \cdot, \mathcal{D})$ gathers elements along $\mathcal{D}$ using $\arg\max_{\mathcal{D}} \mathcal{A}_0$ as indices, with $\mathcal{O} = (\epsilon_{id}, \epsilon_{\text{semantic}})$ as the candidate set.

\textit{
Notably, vs. other fusion approaches, IdAF dispenses with manual masking for target ID fusion region localization, making it more concise and efficient.}

\begin{figure*}[t]
  \centering
   \includegraphics[width=\linewidth]{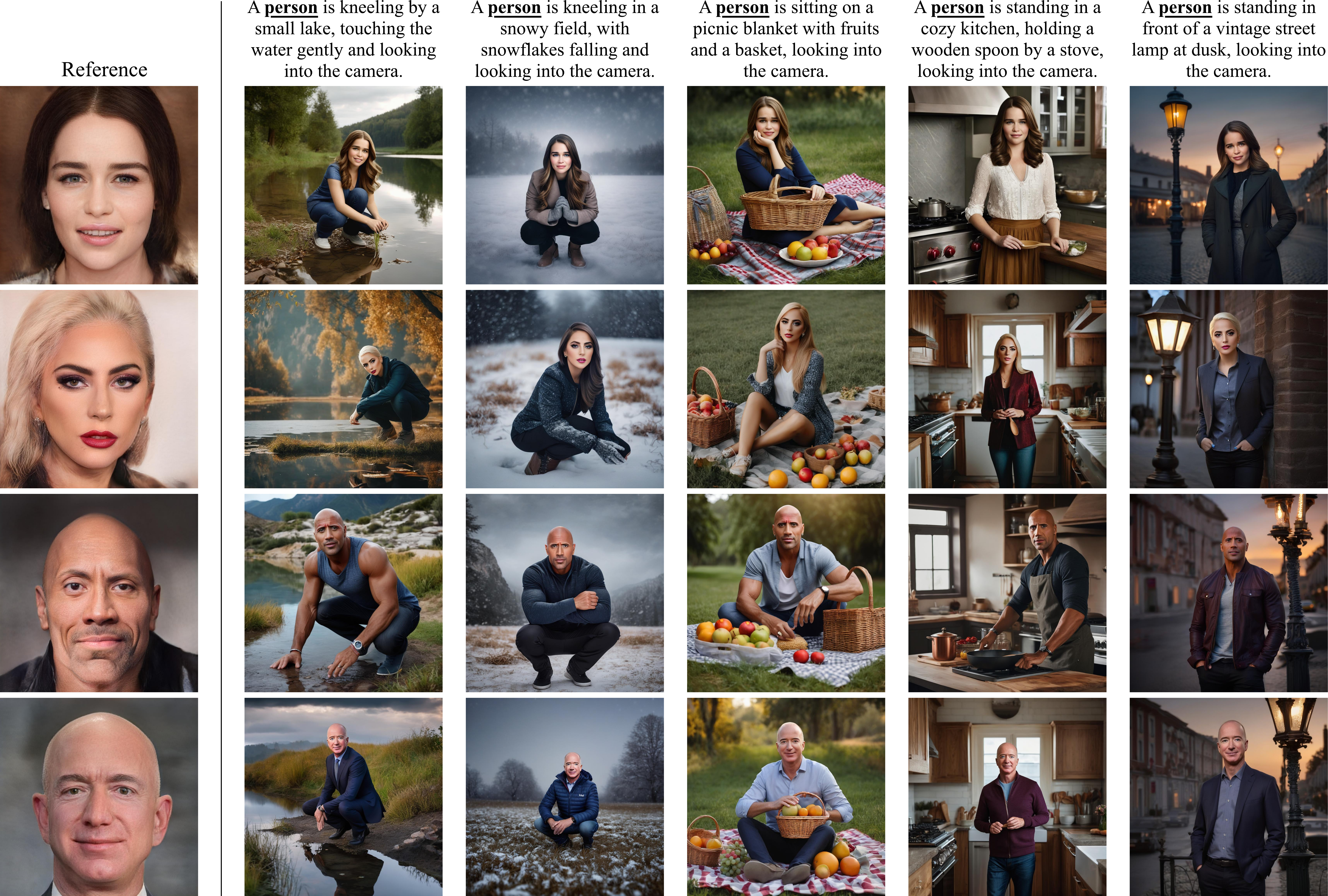}
   \caption{Qualitative Results: Personalized Generation Beyond Facial Close-Ups. 
Our method faithfully renders both character actions and prompt-specified semantic scenes. Generated characters exhibit natural poses and high identity fidelity to reference images, while scenes are aesthetically consistent and visually compelling. This demonstrates significant potential for film and television production applications.}
   \label{fig:main_results}
\end{figure*}

\subsection{Identity Aggregation Prepending (IdAP)}  
In the Identity Adaptive Fusion (IdAF) strategy, semantic scene prompts include phrases like ``a person" to guide character inclusion. Without explicit reference images, however, models generate random identities, causing interference between random and target IDs. To mitigate this, we aim to align random identities with the target ID prior to fusion, and thus design the IdAP processor—suppressing random ID interference via an attention aggregation module and identity prepending.

Taking PhotoMaker~\cite{PhotoMaker} as an example: while the prompt ``a person img" is short, ID insertion via its image encoder expands the embedded token sequence to length 77, with ID information scattered across tokens. To achieve ID embedding without compromising semantic expression, we aggregate these ID tokens into a compact sequence of length $K$. For input ID features $\mathcal{T}_{id} \in \mathbb{R}^{B \times N \times D}$ (where $B$ = batch size, $N$ = sequence length, $D$ = feature dimension), we define a learnable query vector $\mathbf{Q} = \text{Parameter}\left( \mathcal{N}(0, 1; K \times D) \right) \in \mathbb{R}^{K \times D}$ (initialized from a standard normal distribution). Attention scores are computed as:  
\begin{equation}
    \mathbf{S} = \mathcal{T}_{id} \mathbf{Q}^\top \in \mathbb{R}^{B \times N \times K}.
\end{equation}  
Softmax-normalizing $\mathbf{S}$ over the sequence dimension ($\text{dim}=1$) yields attention weights $\mathcal{I} = \text{Softmax}(\mathbf{S}, \text{dim}=1) \in \mathbb{R}^{B \times N \times K}$. Aggregating input features via dimension permutation and matrix multiplication gives:  
\begin{equation}
    \mathcal{T}_{id}^s = \left( \mathcal{I}^{\sigma(0,2,1)} \right) \mathcal{T}_{id} \in \mathbb{R}^{B \times K \times D},
\end{equation}  
where $\sigma(0,2,1)$ denotes dimension permutation.

After obtaining the aggregated ID token $\mathcal{T}_{id}^{s}$, we prepend it to the start of semantic tokens $\mathcal{T}_{\text{semantic}}$ and get final aggregated semantic token $\mathcal{T}_{semantic}^{id}$ (termed ID prepending). Though simple, this step is highly effective: it achieves concatenation (not conflation) of ID and semantic information, preserving the integrity of semantic components. By design, IdAP replaces random identities within  $\mathcal{T}_{semantic}^{id}$ with the prepended target ID information, mitigating random ID interference in the final fused identity.

\noindent \textbf{Explanation on Extensibility.}
We use PhotoMaker~\cite{PhotoMaker} as the baseline, which primarily serves to extract identity features from reference images (i.e., providing identity tokens $\mathcal{T}_{id}$ for our method). Theoretically, our framework can extend the semantic branch to any single-line IPPG method with identity feature extraction capabilities. Integrating IdAF and IdAP as plug-and-play components resolves the model's tendency to generate facial close-ups.

\begin{figure*}[t]
  \centering
   \includegraphics[width=\linewidth]{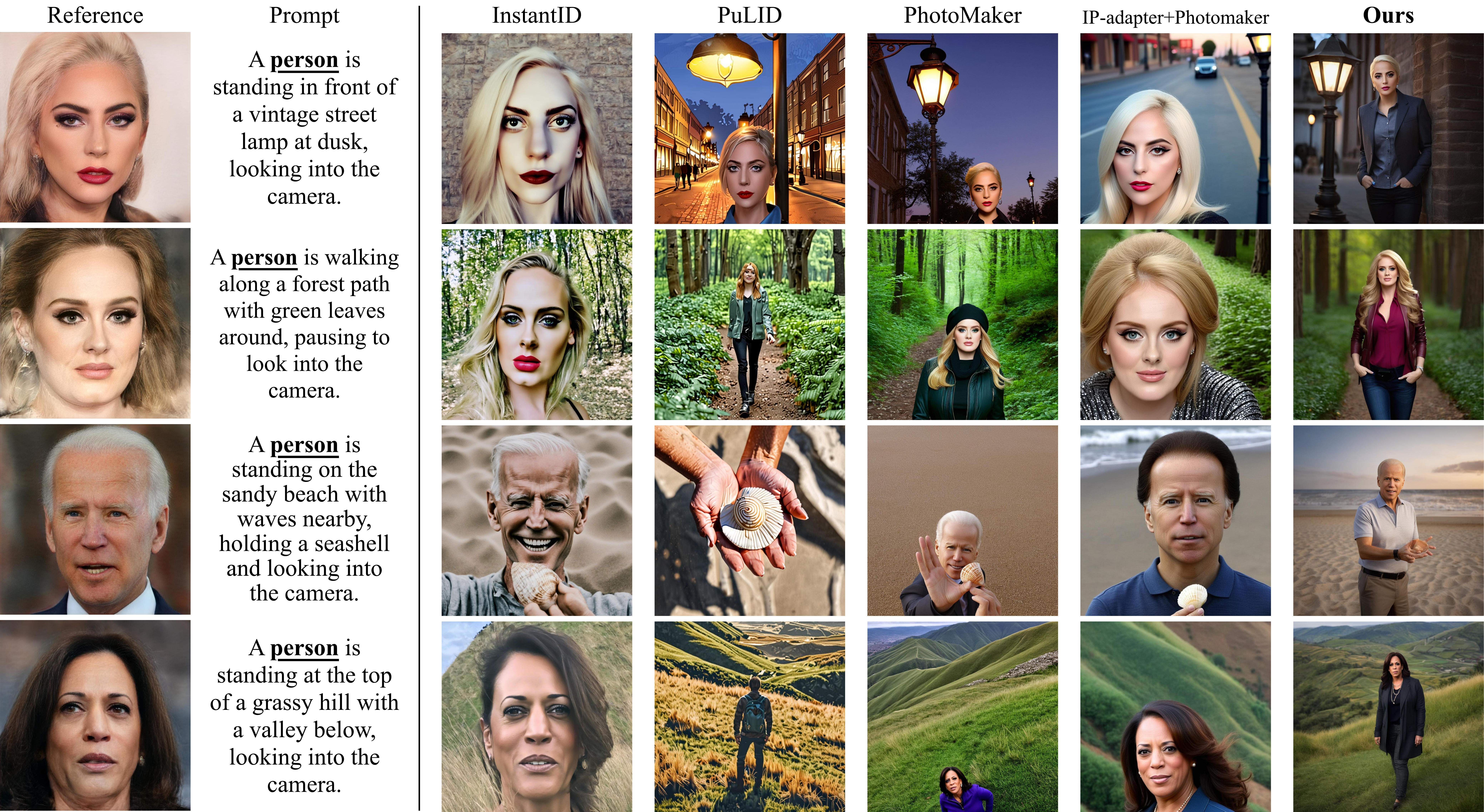}
   \caption{Visual Comparison. Other methods predominantly generate facial close-ups—focusing on heads, lacking full-body poses and actions, with scene semantics only implied through head backgrounds. In contrast, our method naturally renders full-body actions and complete scene contexts, while maintaining high identity fidelity and strong visual appeal.}
   \label{fig:main_compare}
\end{figure*}

\section{Experiments}

\subsection{Experimental Setup}

\textbf{Implementation Details.}  
Our framework is built on PhotoMaker~\cite{PhotoMaker}, leveraging its pre-trained model to extract base identity embedding features before deploying our proposed identity-semantic separated dual-path inference. Both paths use 50 DDIM steps, with inference accelerated on a single Tesla A100.  Hyperparameters in Eq.\eqref{A_0i} are set as $\lambda^0=1$ and $\lambda^1=5$, weighting semantic and identity attention, respectively. The aggregated token length $K$ in the Attention Aggregation Module is empirically set to 8. IdAF and IdAP are activated for sampling timesteps $t > M_1$ ($M_1=10$) and $t > M_2$ ($M_2=15$), respectively.  
For comparisons, we use open-source implementations from GitHub: PhotoMaker~\cite{PhotoMaker} (V2 model), IPadapter~\cite{ipadapter} (improved version from PhotoMaker’s repository, with \textit{ip\_adapter\_scale}=0.7), and PuLID~\cite{PuLID} (following official inference logic in \texttt{app.py} with mode set to ``fidelity'').

\noindent \textbf{Evaluation Metrics.}  
We employ ID similarity (Face sim.)~\cite{ArcFace} to measure identity alignment between references and generated images, and DINO~\cite{dino} to quantify semantic consistency and visual fidelity. Additionally, CLIP-I and CLIP-T scores~\cite{clip} assess image-level similarity and text-image semantic alignment, respectively. Given our core goal of enabling personalized generation beyond facial close-ups—and since facial close-ups with proper backgrounds do not bias semantic consistency evaluations—\textit{we specifically propose using FID~\cite{fID} computed on a full-body human image dataset to accurately quantify our contributions.}

\noindent \textbf{Evaluation Samples.}  
To showcase our method’s capability for personalized generation beyond facial close-ups, we use prompts that encode rich character details and scene contexts—unlike existing works relying on simplistic prompts (e.g., ``a man in the snow," ``a man wearing a santa hat"). Specifically, to balance bodily actions and scene descriptions, we leveraged large language models (LLMs) to randomly generate 20 prompts (e.g., ``A person is kneeling by a small lake, touching the water gently and looking into the camera"), enabling human-free collaboration between LLMs and large vision models. Full prompts are provided in the supplementary materials.  
For reference subjects, we expanded PhotoMaker’s test identities with domain-common human IDs to construct a test dataset of 12 identities.


\begin{figure}[t]
  \centering
   \includegraphics[width=\linewidth]{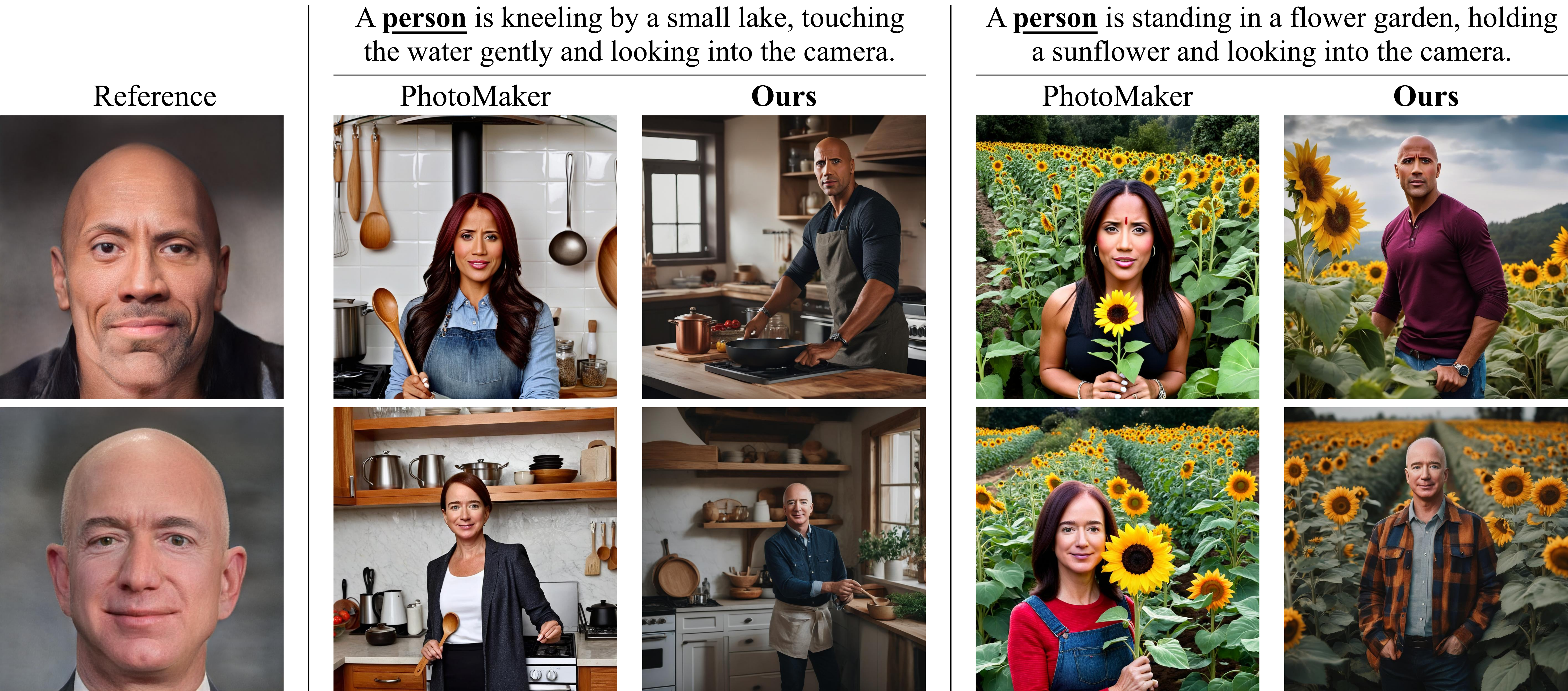}
   \caption{Gender Expression Enhancement. Our method markedly enhances the accuracy of ID gender expression.}
   \label{fig:gender}
\end{figure}

\begin{table*}
\caption{Quantitative Comparative Evaluation Results.}
\label{tab:main_results}
\centering
\begin{tabular}{@{}lccccc@{}}
\toprule
Methods            & Face sim.(\%) $\uparrow$ & DINO(\%) $\uparrow$  & CLIP-I(\%) $\uparrow$ & CLIP-T(\%) $\uparrow$ & \textbf{FID} $\downarrow$    \\ \midrule

PuLID~\cite{PuLID}            & 57.07     & 47.14 & 52.94  & \textbf{32.33}  & 134.03 \\
InstantID~\cite{InstantID}    & 54.68     & \textbf{51.93} & 54.18  & 27.21  & 127.80 \\
IP-Adapter+Photomaker~\cite{ipadapter}&\uline{58.10}    & 47.79  & \textbf{79.28 } & 28.72  & 211.68     \\
Photomaker(Baseline)~\cite{PhotoMaker} & 52.01    & 46.46 & 62.09  & 28.46  & \uline{117.90} \\   
BeyondFacial(Ours) & \textbf{58.67}     & \uline{51.24} & \uline{64.87}  & \uline{31.47}  & \textbf{54.62}  \\ \midrule  
Imp.(Baseline)$\uparrow$ & 6.67   & 4.78 & 2.79  & 3.01  & 63.28  \\ 
\bottomrule
\end{tabular}
\end{table*}

\subsection{Qualitative Evaluation}

Based on the aforementioned evaluation prompts and test identities, we conducted qualitative assessments. Fig.~\ref{fig:main_results} presents generated results for selected test identities across 5 prompts, with additional visualizations provided in the supplementary materials.  
Experimental results confirm that our method faithfully renders both character actions and prompt-specified semantic scenes—addressing the core limitation of existing methods (over-reliance on facial close-ups). Generated characters exhibit natural poses and high identity fidelity to references, while scenes are aesthetically consistent and visually compelling. This demonstrates our framework’s unique ability to enable high-quality personalized generation beyond facial close-ups, with significant potential for film and television production applications.

\noindent \textbf{Gender Expression Enhancement. }  
Existing personalized generation methods typically necessitate manual replacement of gender-specific placeholders to align with reference subjects’ gender. For gender-agnostic prompts (e.g., ``A person"), generated images often exhibit gender ambiguity. As shown in Fig.~\ref{fig:gender}, our method not only addresses the facial close-up bias but also markedly improves gender expression accuracy for target identities—bolstering overall identity fidelity. This ancillary benefit stems from our IdAP module, which precisely aligns semantic tokens with target identity features, highlighting the framework’s capacity to refine fine-grained identity attributes beyond core identity preservation.

\begin{figure}[t]
  \centering
   \includegraphics[width=\linewidth]{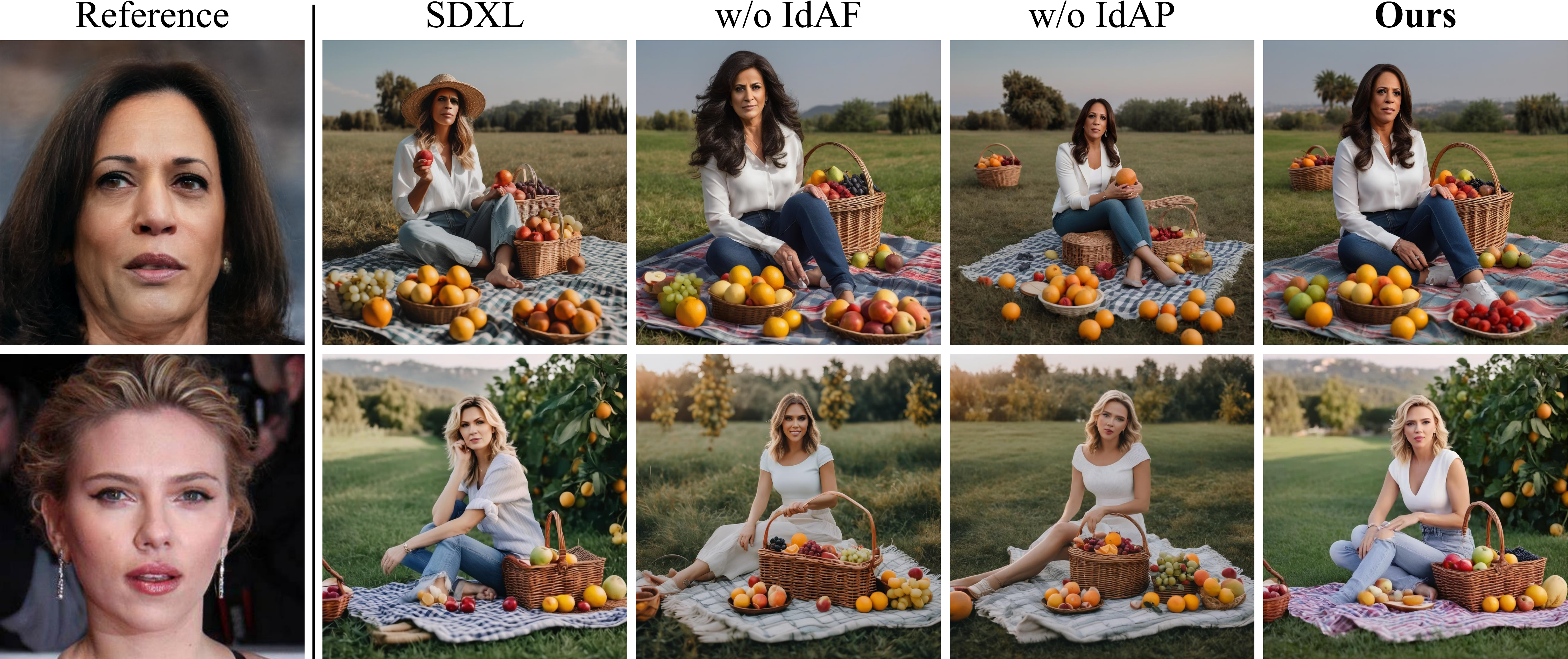}
   \caption{Visualization Results of Module Ablation Studies. Sequential use of IdAF and IdAP steadily enhances target ID similarity—without disrupting original semantics or causing issues like facial close-ups.}
   
   \label{fig:ablation_Module}
\end{figure}

\begin{table}
\caption{Quantitative Results of Module Ablation Studies.}
\label{tab:ablation}
\centering
\begin{tabular}{@{}lcccc@{}}
\toprule
Methods            & Face sim. & DINO  & CLIP-I& CLIP-T    \\ 
           & (\%) $\uparrow$ & (\%) $\uparrow$  & (\%) $\uparrow$ & (\%) $\uparrow$    \\ \midrule

SDXL~\cite{stable_diffusion}  & - &13.37 &55.78& 31.00\\
Ours(w/o IdAF)           & 36.34     & 45.58 & 57.27  & 30.97   \\
Ours(w/o IdAP)           & 54.17     & 37.37 & 56.28  & 31.46  \\
Ours(Full) & 58.67     & 51.24 & 64.87  & 31.47  \\ 
\bottomrule
\end{tabular}
\end{table}

\begin{table*}[t]
\centering
\setcounter{table}{2}
\caption{User Study. Our method demonstrates significant advantages across all four evaluation dimensions.}
\vspace{-0.3cm}
\label{tab:user_study}
\begin{tabular}{lccccc}
\hline
\multicolumn{1}{c}{{\color[HTML]{000000} }}  & {\color[HTML]{000000} Identity similarity}                    & {\color[HTML]{000000} \begin{tabular}[c]{@{}c@{}}Image-text \\ alignment \end{tabular}}                  & {\color[HTML]{000000} \begin{tabular}[c]{@{}c@{}}Image quality \& \\ visual aesthetics\end{tabular}} & {\color[HTML]{000000} \begin{tabular}[c]{@{}c@{}}Scene construction \\ \& figure integrity \end{tabular}} & AVREGE\\ \hline
{\color[HTML]{000000} InstantID~\cite{InstantID}}             & {\color[HTML]{000000} 50.89}                                  & {\color[HTML]{000000} 48.06}                                  & {\color[HTML]{000000} 47.20}                                                                           & {\color[HTML]{000000} 49.35}        &  {\color[HTML]{000000} 48.88}                                                                  \\
{\color[HTML]{000000} PuLID~\cite{PuLID}}                 & {\color[HTML]{000000} 57.21}                                  & {\color[HTML]{000000} 70.95}                                  & {\color[HTML]{000000} 61.99}                                                                          & {\color[HTML]{000000} 68.68}          &    {\color[HTML]{000000} 64.71}                                                              \\
{\color[HTML]{000000} PhotoMaker~\cite{PhotoMaker}}            & {\color[HTML]{000000} 47.95}                                  & {\color[HTML]{000000} 60.85}                                  & {\color[HTML]{000000} 56.89}                                                                          & {\color[HTML]{000000} 60.50}        &   {\color[HTML]{000000} 56.55}                                                                  \\
{\color[HTML]{000000} IP-Adapter+Photomaker~\cite{ipadapter}} & {\color[HTML]{000000} 63.58}                                  & {\color[HTML]{000000} 48.45}                                  & {\color[HTML]{000000} 53.35}                                                                          & {\color[HTML]{000000} 52.51}        &    {\color[HTML]{000000} 54.47}                                                                \\
{\color[HTML]{000000} \textbf{Ours}}         & \cellcolor[HTML]{FDE6E5}{\color[HTML]{000000} \textbf{75.17}} & \cellcolor[HTML]{FDE6E5}{\color[HTML]{000000} \textbf{79.01}} & \cellcolor[HTML]{FDE6E5}{\color[HTML]{000000} \textbf{74.38}}                                         & \cellcolor[HTML]{FDE6E5}{\color[HTML]{000000} \textbf{78.34}}          &      \cellcolor[HTML]{FDE6E5}{\color[HTML]{000000} \textbf{76.73}}                            \\ \hline
\end{tabular}
\end{table*}
\begin{figure*}[t]
  \centering
\includegraphics[width=\linewidth]{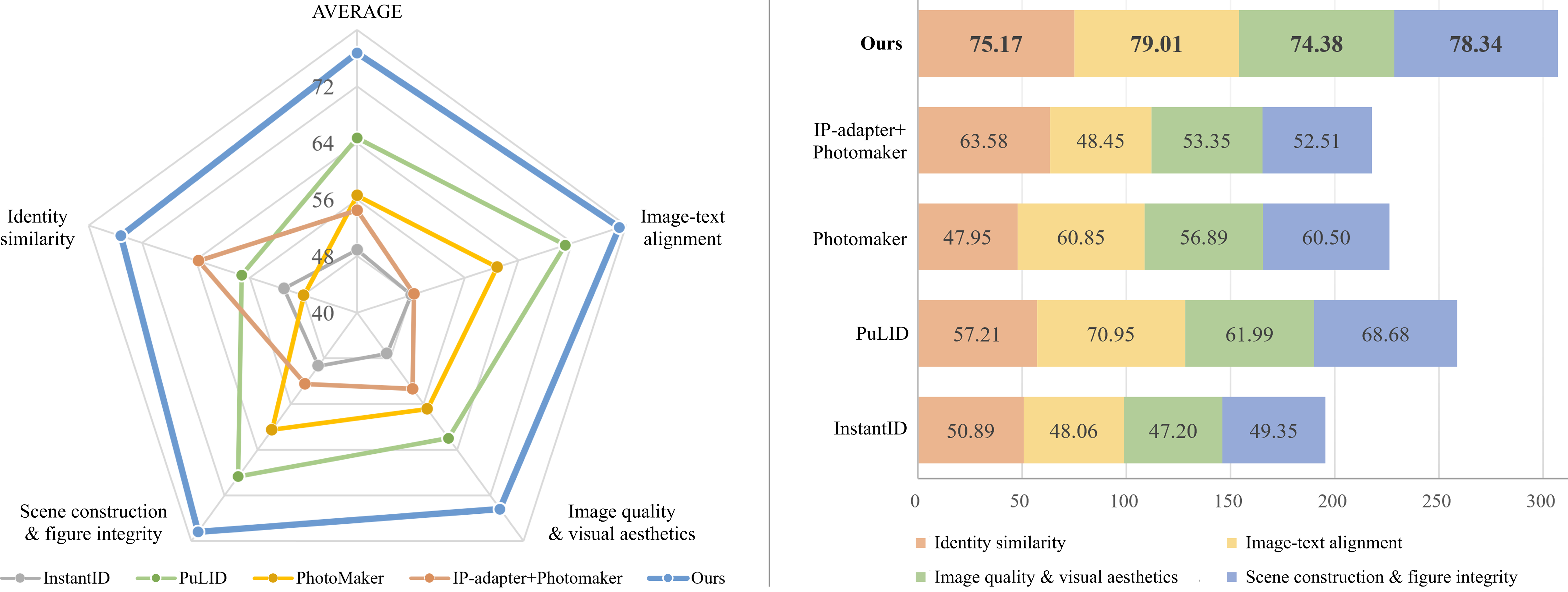}
\vspace{-0.5cm}	
   \caption{The visualization of our user study results—presented in the form of a radar chart and a stacked bar chart—clearly illustrates the significant advantages of our method.}
   \vspace{-0.5cm}	
   \label{fig:user_study}
\end{figure*}

\subsection{Comparison and Analysis}

\textbf{Visual Comparison.}  
Given reference identities and prompts, we compare our method with state-of-the-art works (InstantID~\cite{InstantID}, PhotoMaker~\cite{PhotoMaker}, IP-Adapter+PhotoMaker~\cite{ipadapter}, PuLID~\cite{PuLID}). Selected results are shown in Fig.~\ref{fig:main_results}, with additional comparisons in the supplementary materials.  
Visualizations reveal critical limitations of existing methods: (1) InstantID and IP-Adapter+PhotoMaker suffer from the inherent facial close-up issue—their outputs focus on head regions, lack full-body poses/actions, and relegate scene semantics to mere background replacement behind the head; (2) while PhotoMaker reduces the head’s proportion in the frame, the character remains positioned at the image edge, failing to display a complete body or intended actions; (3) PuLID fails under long action+scene prompts, leading to character loss or missing facial details.
In stark contrast, our method resolves these limitations by simultaneously rendering full-body actions and complete scene contexts. It achieves natural character movements, high identity fidelity, and strong visual appeal—directly addressing the core challenge of personalized generation beyond facial close-ups.

\noindent \textbf{Quantitative Comparison.}  

For quantitative evaluation, we curated a 4,800-sample dataset (12 identities × 20 prompts × 20 seeds) for each method. Since our full-body outputs reduce facial proportions (complicating identity similarity assessment), we used expression-emphasized prompts for identity metrics to ensure facial prominence. For FID, we constructed a 4,800-sample SDXL dataset~\cite{stable_diffusion} of full-body, action-rich scenes with random identities—tailored to reflect our focus on generation beyond close-ups.  

Results in Table \ref{tab:main_results} show our method achieves top/second-top performance across all metrics(Bold: optimal values; Underlined: suboptimal values), balancing identity fidelity and semantic consistency. Superior Face sim. validates IdAP’s role in identity preservation. Improved DINO, CLIP-I, and CLIP-T over PhotoMaker confirm IdAF prevents identity embeddings from disrupting scenes. Most notably, our FID improves several-fold over baselines, confirming robust film-grade character-scene generation capability and strong practical potential.

\subsection{User Study }
Given the core goal of this work—identity-preserving personalized generation beyond facial close-ups with complete human figures and rich scene semantics—we prioritize not only identity similarity (Face Sim.~\cite{ArcFace}) but also the integrity of generated humans/scenes and visual aesthetics, distinct from metrics like DINO~\cite{dino} and CLIP-I~\cite{clip}. In the main text, we compute FID~\cite{fID} on a dataset scaled to match SDXL-generated~\cite{stable_diffusion} outputs (complete humans, rich scenes), effectively evaluating our method’s substantive progress toward this goal. As shown in Table 1 of the main text, our FID results demonstrate a significant margin over existing works in personalized generation beyond facial close-ups.

To comprehensively assess our method’s advancements, we conduct a user study to gather subjective human evaluations—this study strongly validates practical application potential. Specifically, we set InstantID~\cite{InstantID}, PhotoMaker~\cite{PhotoMaker}, PuLID~\cite{PuLID}, and IP-Adapter+PhotoMaker~\cite{ipadapter} as baselines, presenting 4 generation tasks (each with reference ID-text-generated image triples for 5 methods: 4 baselines + ours) to each participant. Respondents score 20 combinations (4 tasks × 5 methods) across four criteria using a 100-point scale (minimum increment of 5 points).
\begin{itemize}[leftmargin=*, noitemsep, topsep=2pt]
\item \textbf{Identity similarity} (generated vs. reference figures): Assesses identity consistency.
\item \textbf{Image-text alignment} (generated content vs. text prompts) : Assesses cross-modal consistency.
\item \textbf{Image quality \& visual aesthetics}: Assesses perceptual quality and aesthetic appeal.
\item \textbf{Scene construction \& figure integrity}: Assesses completeness of scenes and human figures.
\end{itemize}
Specifically, we anonymized all method names in the questionnaire and randomly shuffled the method order for each response set. Statistical results are based on feedback from 100 participants, with details presented in Table \ref{tab:user_study}.

The user study results demonstrate our method’s significant advantages over all baselines across the four evaluation criteria (identity similarity, image-text alignment, image quality \& visual aesthetics, scene construction \& figure integrity), with consistent user approval. \textbf{This validates the success of our identity-preserving personalized generation (IPPG) beyond facial close-ups.}
For absolute scores, our method achieves the highest performance in image-text alignment (79.01) and scene construction \& figure integrity (78.34). For relative gains, it outperforms the second-ranked method by the largest margins in identity similarity (11.59) and image quality \& visual aesthetics (12.39).
PuLID~\cite{PuLID} ranks second in overall average (64.71) but suffers a notable deficit in identity similarity (57.21), indicating substantial degradation of its ID fidelity with increasing semantic complexity. Equipped with dual ID enhancement, IP-Adapter+PhotoMaker~\cite{ipadapter} achieves the second-highest identity similarity (63.58)—surpassing other baselines and trailing only ours—yet attains the lowest image-text alignment. This further confirms that \textbf{balancing ID fidelity and image-text alignment remains a key challenge in contemporary IPPG research.}

In contrast to baselines, \textbf{our method delivers substantial gains across all criteria and achieves a meaningful balance between ID fidelity and cross-modal consistency—addressing the core tradeoff in personalized generation beyond facial close-ups.} Fig.\ref{fig:user_study} visualizes the results of our user study in the form of a radar chart and a stacked bar chart, clearly demonstrating the significant advantages of our method.

\begin{figure}[t]
  \centering
   \includegraphics[width=\linewidth]{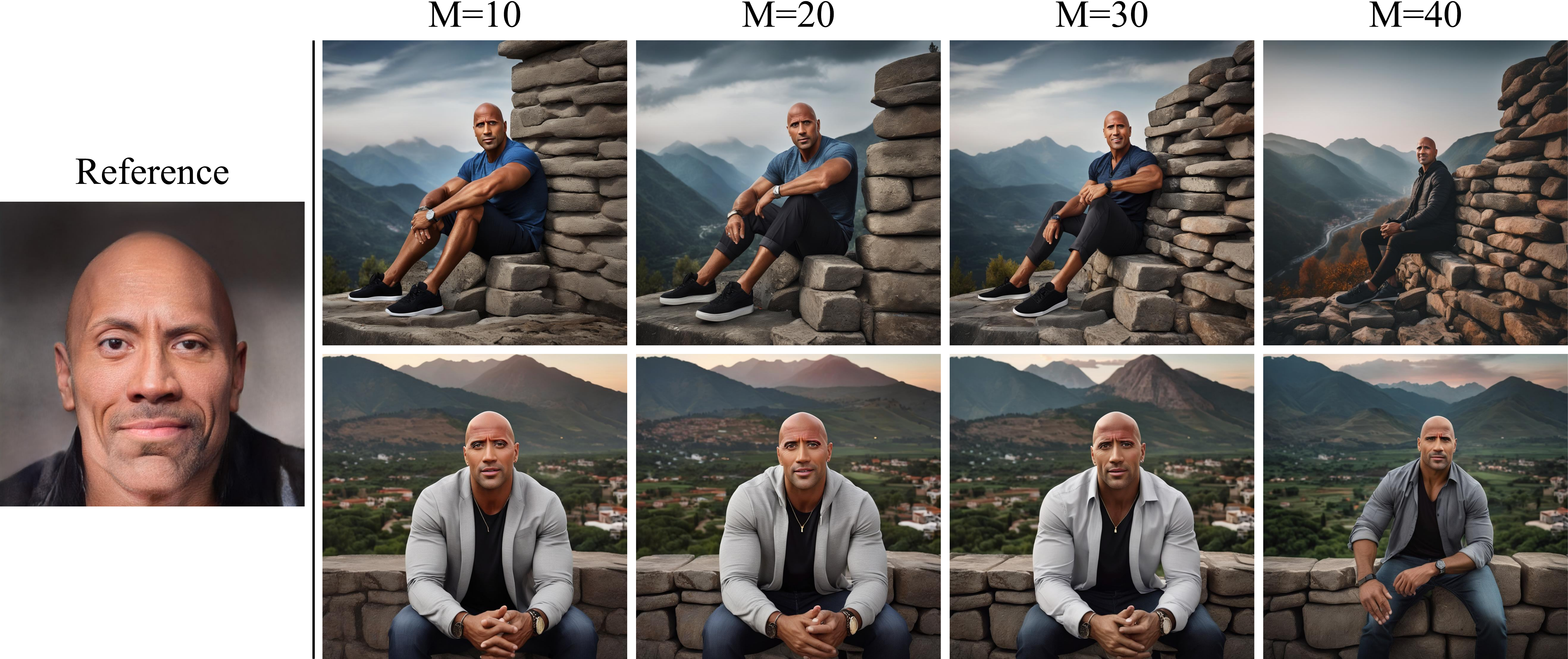}
   \caption{Exploration of IdAF Activated Timesteps. ID fidelity decreases with delayed fusion timing.}
   \label{fig:ablation_step}
\end{figure}

\begin{figure}[t]
  \centering
  \includegraphics[width=\linewidth]{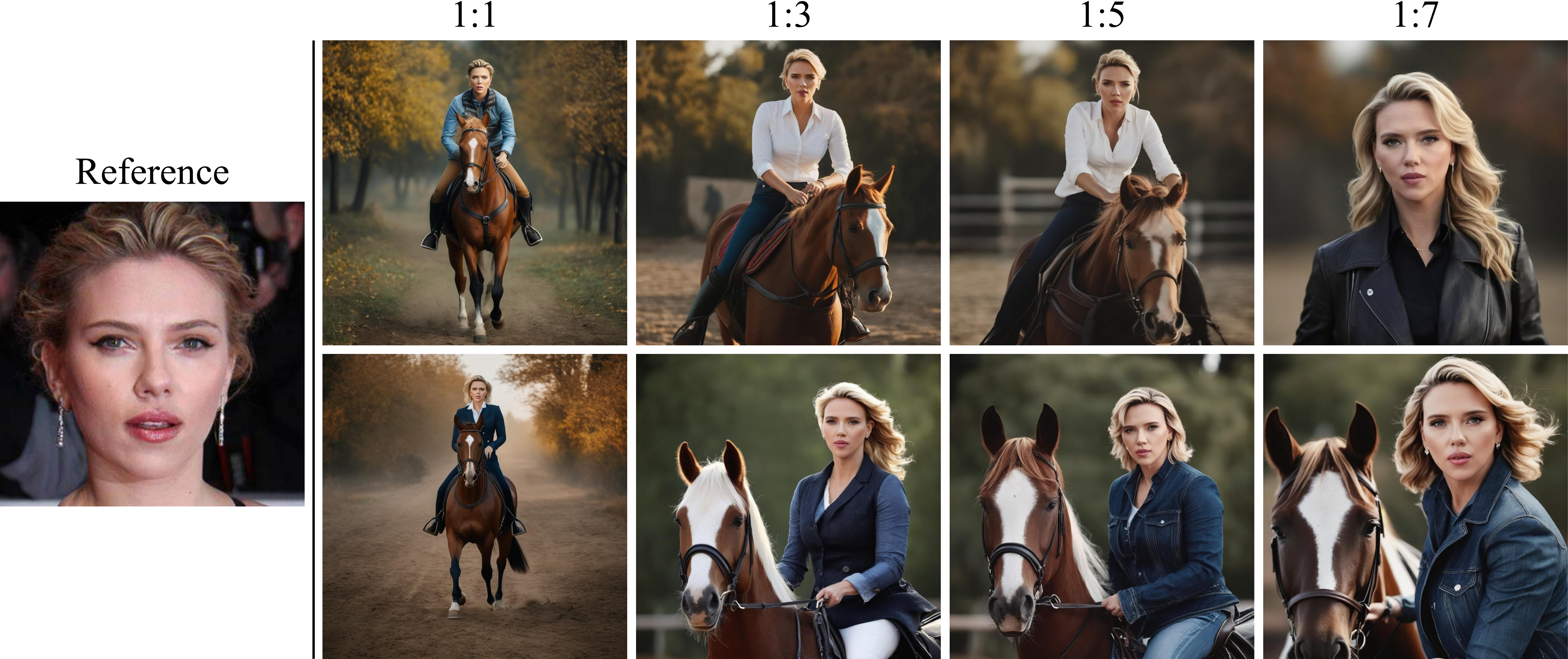}
   \caption{The Impact of Attention Weight Scaling Factor on Facial Close-ups. As the proportion of ID attention weights (the latter in the ratio) increases, generated images gradually exhibit a facial close-up tendency.}
   \label{fig:ablation_lambda}
\end{figure}

\subsection{Ablation Study}

\textbf{Module Contributions.}   
Ablation studies validate IdAF and IdAP’s independent contributions (Table \ref{tab:ablation}). Our method uses SDXL as the baseline (leveraging its scene layout, with random identities precluding Face sim. computation).  
Key findings: IdAP primarily drives ID similarity (Face sim.) gains, while IdAF more strongly enhances semantic consistency (DINO). CLIP-I improves notably only when both modules are combined (synergistic effect), and CLIP-T remains stable (slightly exceeding SDXL with IdAP), confirming preserved textual alignment.  Visualizations in Fig.\ref{fig:ablation_Module} confirm sequential application of IdAF and IdAP progressively boosts target identity similarity, preserves semantics, and avoids facial close-up biases—validating their complementary roles in balancing identity fidelity and scene coherence.

\begin{figure*}[t]
  \centering
   \includegraphics[width=\linewidth]{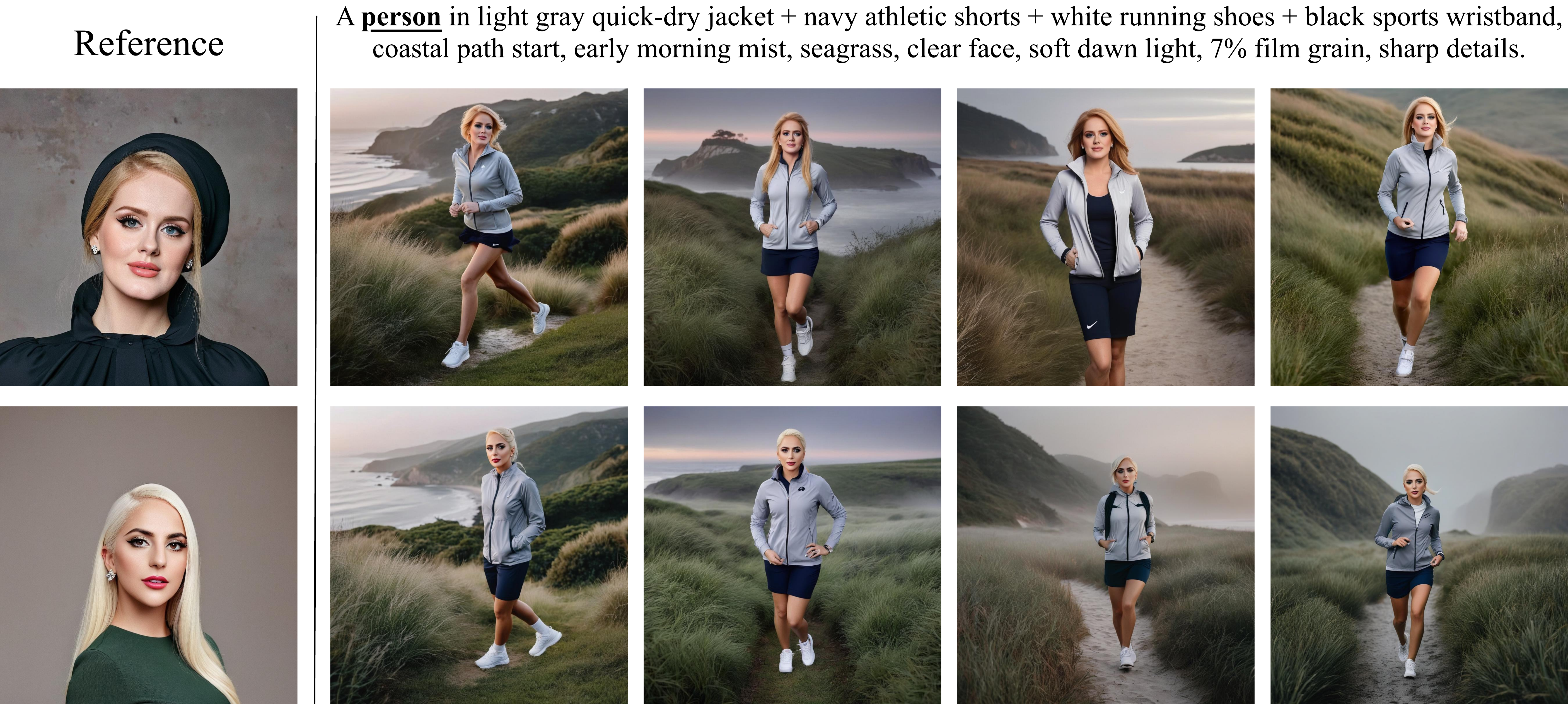}
   \vspace{-0.5cm}	
   \caption{Application Extension: Consistent action sequence generation. Under the same semantic guidance, BeyondFacial enables action sequence generation for specific characters, incorporating consistency in scenes and costumes.}
   \vspace{-0.2cm}	
   \label{fig:Action_sequence}
\end{figure*}

\begin{figure*}[t]
  \centering
   \includegraphics[width=\linewidth]{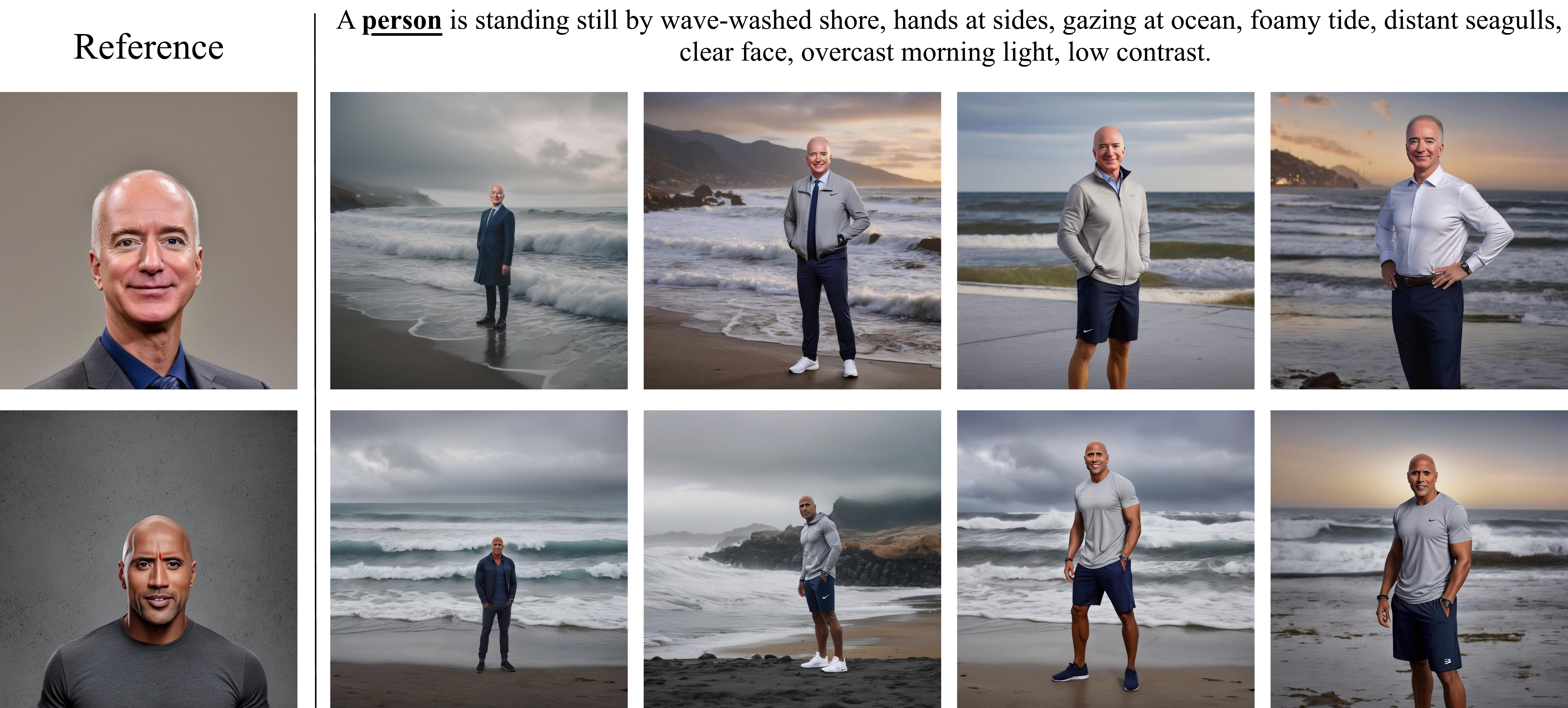}
   \vspace{-0.5cm}	
   \caption{Application Extension: Multi-scale(zoomable) shot genertion. Under the same semantic guidance, BeyondFacial enables multi-scale shot generation for specific characters—analogous to zoomable lens photography—and facilitates character generation with far-to-near transitions.}
   \vspace{-0.5cm}	
   \label{fig:Muti_shot}
\end{figure*}

\begin{figure*}[t]
  \centering
   \includegraphics[width=\linewidth]{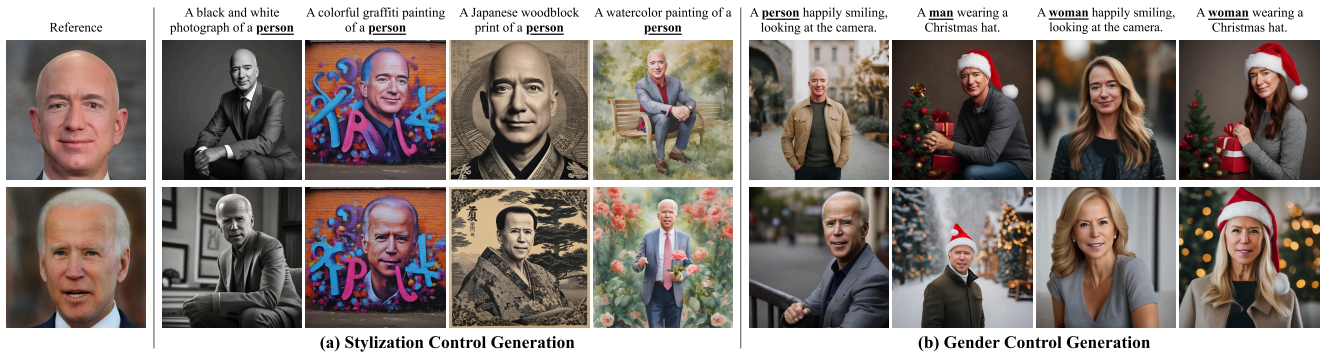}
   \caption{Applications. Our method enables film-grade character-scene generation beyond close-ups, with extended capabilities: stylization and gender control. }
   \label{fig:Style_gender}
\end{figure*}

\noindent \textbf{Attention Weight Scaling Factor. }  
We performed ablation analysis on the attention weight scaling factors $\lambda^{(i)}(i=0,1)$ in the IdAF strategy (Eq. \eqref{eq:A0i}), which are key to fusing identity and semantic scene information. As shown in Fig. \ref{fig:ablation_lambda}, we tested ratios of semantic to identity attention weights ($\lambda^{(0)}:\lambda^{(1)}$) of 1:1, 1:3, 1:5, and 1:7 across two random seeds. 
Results reveal that increasing the proportion of identity attention weights leads to generated images increasingly exhibiting a facial close-up bias—directly validating the pivotal role of our IdAF strategy in mitigating this issue. At a 1:1 ratio, facial regions were excessively small, hindering intuitive identity fidelity assessment. Thus, we uniformly adopted a 1:5 weight ratio in all experiments.

\noindent \textbf{IdAF Activated Timesteps. }  
Timing is critical for IdAF to fuse identity and semantics without interference: early fusion (before semantic layouts stabilize) impairs scene rendering, while late fusion (after identity features solidify) degrades identity fidelity.  
With total DDIM steps $T=50$ ($t:0 \to T$) and IdAF activated for $t>M$, we tested $M=10,20,30,40$ (Fig. \ref{fig:ablation_step}). Results confirm identity consistency degrades progressively for $M=30,40$. Balancing robustness and performance, we fix $M=15$ in all experiments.

\subsection{Applications}  
\subsubsection{Cinematic-Grade Character-Scene  generation}
To validate our method BeyondFacial’s capability for Cinematic-Grade Character-Scene(CGCS) generation, we first use a large language model (LLM) to generate character-centric story narratives with action and scene transitions. The generated narratives are presented below:

\textit{``A person embraces the quiet of early morning along a windswept coastal path, seeking calm amid the rhythm of waves. Dressed in a light gray quick-dry sports jacket, navy blue athletic shorts, white running shoes, and a black silicone sports wristband on the left wrist, their outfit is practical yet sleek—perfect for the crisp sea air. The story unfolds as they move from the path’s start to a cliffside overlook, pausing to absorb the dawn, with salt breezes rustling their hair and golden light gradually warming the horizon."
}

\begin{figure*}[t]
  \centering
   \includegraphics[width=\linewidth]{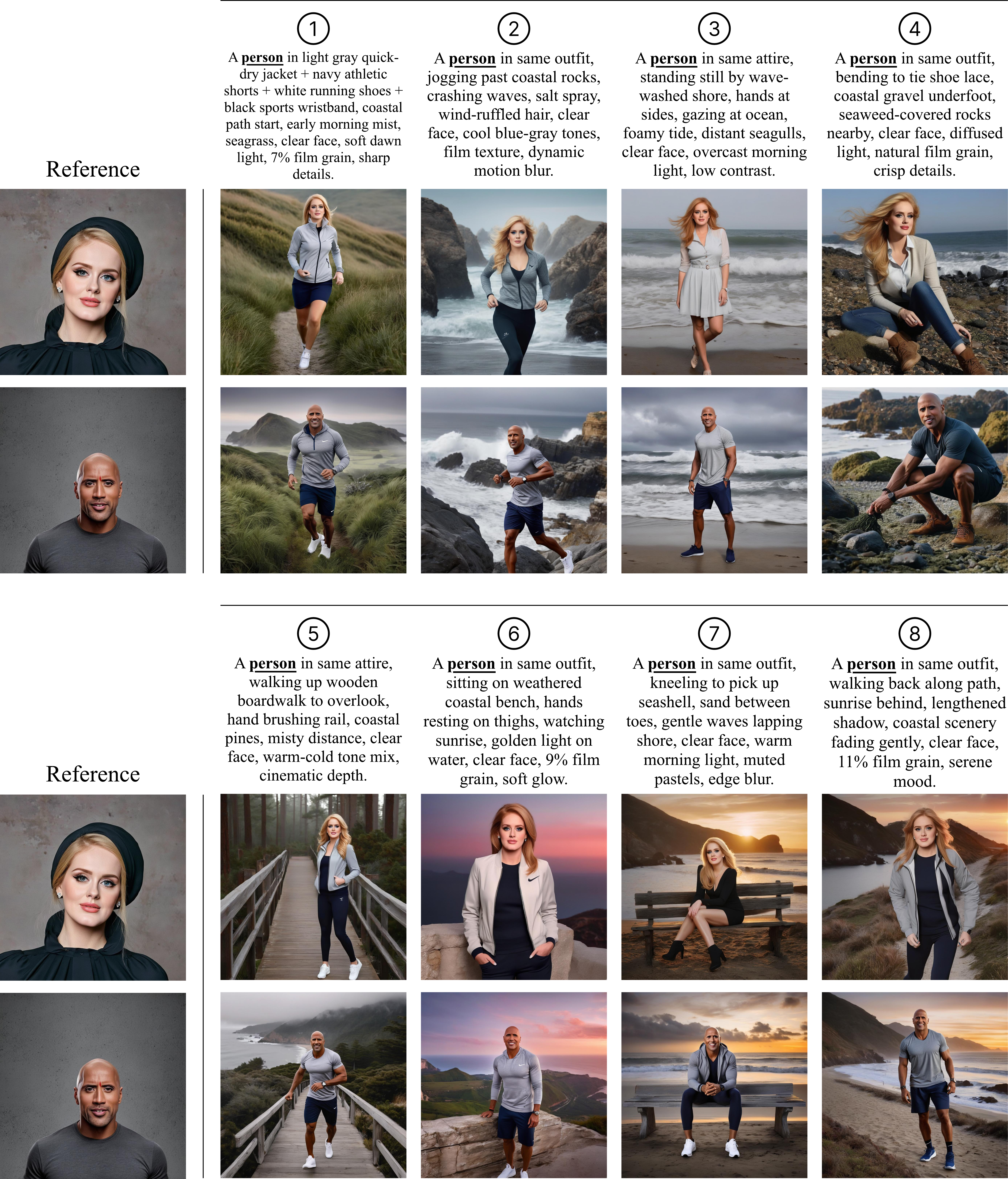}
   \caption{Application Extension: Storyboard Scene Generation Results Based on BeyondFacial (1). Our method achieves personalized generation that balances ID fidelity and semantic expressiveness—delivering complete human figures and visually compelling cinematic scenes.}
   \label{fig:Movie_1}
\end{figure*}

\begin{figure*}[t]
  \centering
   \includegraphics[width=\linewidth]{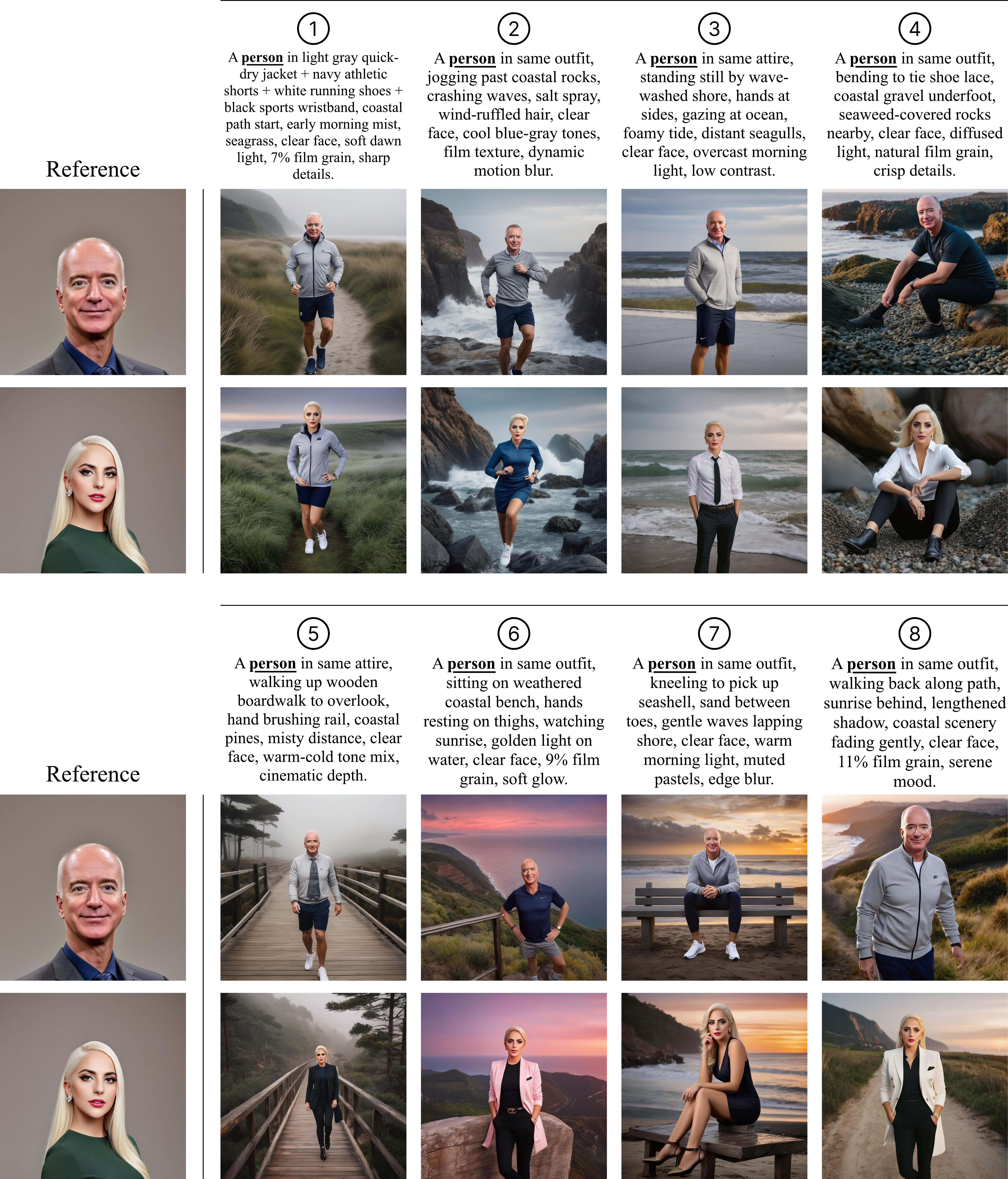}
   \caption{Application Extension: Storyboard Scene Generation Results Based on BeyondFacial (2). Our method achieves personalized generation that balances ID fidelity and semantic expressiveness—delivering complete human figures and visually compelling cinematic scenes.}
   \label{fig:Movie_2}
\end{figure*}

\begin{figure*}[t]
  \centering
   \includegraphics[width=\linewidth]{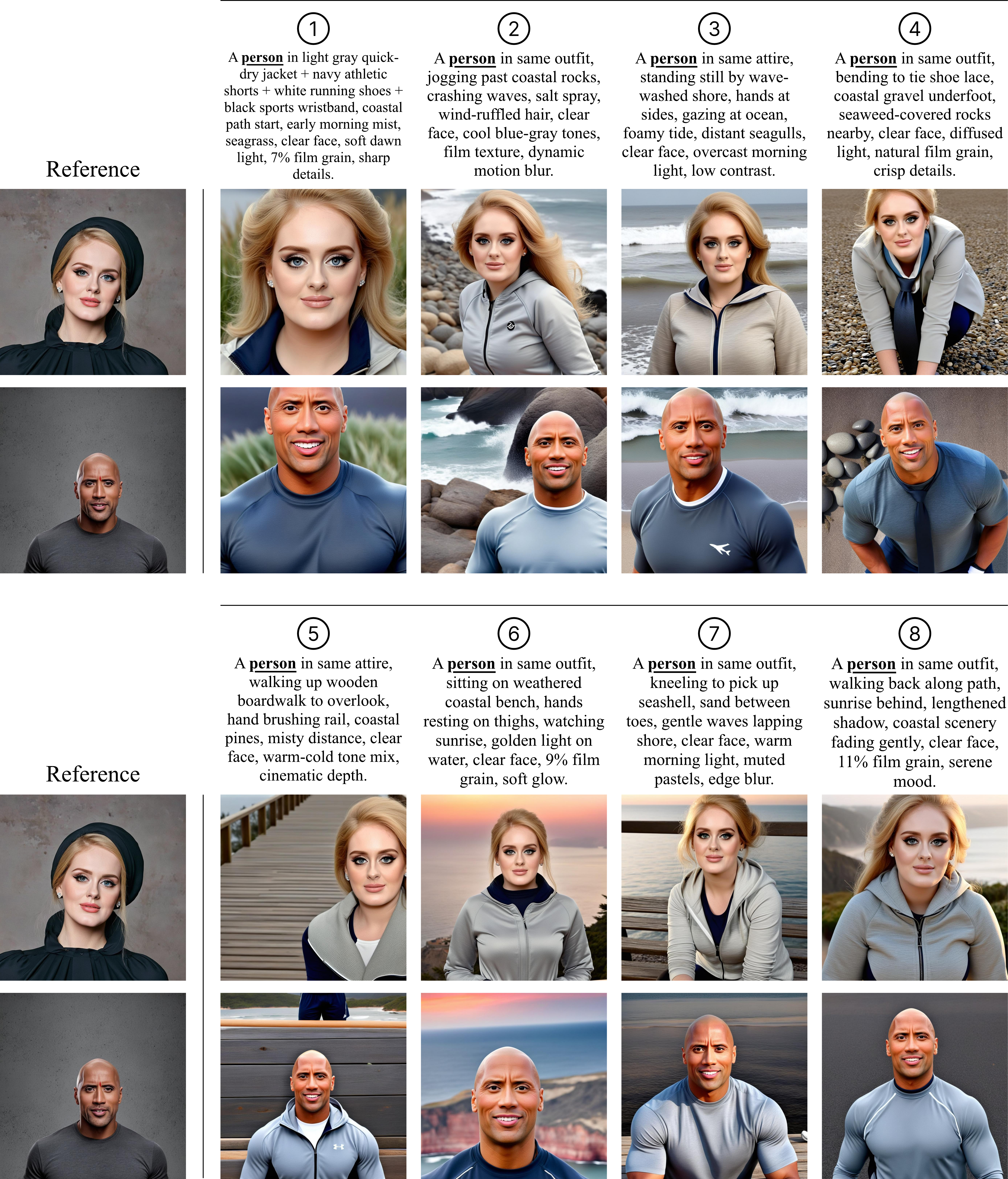}
   \caption{Application Extension: Storyboard Scene Generation Results Based on IP-Adapter+Photomaker~\cite{ipadapter}. IP-Adapter+PhotoMaker’s uniform facial close-ups highlight a notable deficit in character-scene generation.}
   \label{fig:Movie_1_Compare}
\end{figure*}

Subsequently, we prompt the large language model (LLM) to generate 8 consecutive shot-specific scene descriptions from this narrative, serving as prompts. For a given reference ID, our BeyondFacial method generates 8 character-scene images featuring the reference ID as the core character. 
Fig.\ref{fig:Movie_1} and \ref{fig:Movie_2} visualize our method’s results for four test IDs. For LLM-generated cinematic shot prompts (complex character-action-scene descriptions),\textbf{ our method achieves personalized generation that balances ID fidelity and semantic expressiveness—delivering complete human figures and visually compelling cinematic scenes}. For intuitive comparison, we include results from IP-Adapter+PhotoMaker~\cite{ipadapter} (Fig.\ref{fig:Movie_1_Compare}): \textbf{its uniform facial close-ups highlight the gap in CGCS generation capability, underscoring our method’s unique value in beyond-facial personalized creation.}

Furthermore, we observe from experimental results that BeyondFacial supports \textbf{consistent action sequence generation} (Fig.\ref{fig:Action_sequence}) and \textbf{multi-scale (zoomable) shot generation} (Fig.\ref{fig:Muti_shot}). This delivers substantial value to IPPG’s application in CGCS generation.

\subsubsection{Style and Gender Control Generation}
Beyond generating film-grade character-scene content (surpassing the facial close-up limitations of conventional IPPG methods), our framework supports extended capabilities in stylization and gender control. Fig.~\ref{fig:Style_gender} illustrates these functionalities: left panels showcase stylization control, while right panels demonstrate gender control. For gender-agnostic prompts (e.g., ``A person"), our method consistently aligns with the inherent gender of reference identities. Notably, to enable intuitive evaluation of identity fidelity and control performance, we omitted panoramic cues (e.g., ``full body") from text inputs—validating the robustness of our approach without explicit structural prompts.


\section{Conclusion}  
To address the prevalent facial close-up limitation in existing Identity-Preserving Personalized Generation (IPPG) methods, this work presents an innovative framework for personalized generation beyond facial close-ups. It achieves a balanced integration of identity fidelity and semantic scene rendering, enabling film-grade character-scene image generation with significant practical value.  
Specifically, we design a dual-line inference pipeline (DLI) to decouple identity and semantic scene reasoning. Building on DLI, we introduce the Identity Adaptive Fusion (IdAF) strategy for interference-free identity-semantic fusion and the Identity Aggregation Prepending (IdAP) module to further enhance identity fidelity.  
Extensive quantitative and qualitative evaluations, alongside comparative analyses, validate the effectiveness and robustness of our method in personalized generation beyond facial close-ups. Our approach is convenient (no manual masking required) and efficient (tuning-free), with substantial potential for film and television production. Moreover, as a plug-and-play component, it can be readily integrated into existing IPPG frameworks to mitigate their facial close-up biases.

\clearpage
{
    \small
    \bibliographystyle{ieeenat_fullname}
    \bibliography{DM_NAF}
}


\end{document}